\documentclass[11pt,a4paper]{article}
\usepackage[hyperref]{acl2021}
\usepackage{times}
\usepackage{latexsym}
\usepackage{graphicx}
\usepackage[toc,page]{appendix}

\usepackage{microtype}

\aclfinalcopy 
\def\aclpaperid{886} 

\title{Context-gloss Augmentation for Improving Word Sense Disambiguation}

\author{Guan-Ting Lin \\
 National Tsing-Hua University \\
 \texttt{daniel094144@gmail.com} \\\And
 Manuel Giambi \\
 National Tsing-Hua University \\
 \texttt{giambi.manuel@gmail.com} \\}

\date{January 2021}

\begin{document}
\maketitle
\begin{abstract}
The goal of Word Sense Disambiguation (WSD) is to identify the sense of a polysemous word in a specific context. Deep-learning techniques using BERT have achieved very promising results in the field and different methods have been proposed to integrate structured knowledge to enhance performance. At the same time, an increasing number of data augmentation techniques have been proven to be useful for NLP tasks. Building upon previous works leveraging BERT and WordNet knowledge, we explore different data augmentation techniques on context-gloss pairs to improve the performance of WSD. In our experiment, we show that both sentence-level and word-level augmentation methods are effective strategies for WSD. Also, we find out that performance can be improved by adding hypernyms' glosses obtained from a lexical knowledge base. We compare and analyze different context-gloss augmentation techniques, and the results show that applying back translation on gloss performs the best.
\end{abstract}

\section{Introduction}
Word Sense Disambiguation (WSD) is a fundamental task and long-standing challenge in Natural Language Processing (NLP), which aims to find the exact sense of an ambiguous word in a particular context \cite{navigli-2009}. Previous approaches to WSD can be grouped into two main categories: knowledge-based and supervised methods.

Knowledge-based WSD methods use lexical databases like WordNet \cite{wordnet}. WordNet provides a definition for each word sense, called gloss. Gloss has been used extensively in WSD, first in Lesk algorithm \cite{Lesk1986AutomaticSD} and subsequently in many other approaches \cite{lesk-2, lesk-3}.


As for supervised methods, traditional approaches \cite{makes-sense, wsd-wiki, embeddings} focus on training many \textit{word expert} classifiers which specialise in disambiguating one lemma each. More recent methods \cite{bi-lstm, neural-wsd} using neural networks aim at improving the performance in the all-word WSD task. \citet{glosses-wsd-2, glosses-wsd} is the first to use glosses during training. Those attempts show the importance of including gloss information into the WSD task.

More recently, \citet{gloss-bert} proposed a new approach named GlossBERT that considerably improves the performance on WSD. They generate concatenated context-gloss pairs using WordNet and then use them to fine-tune a pre-trained BERT \cite{bert} model in a binary classification task. In addition, an increasing number of works \cite{bert-wsd, bi-encoders} adopt two separate BERT models to encode context and gloss respectively. These works achieved competitive results and \citet{bert-wsd} also revealed that adding gloss information can alleviate the poor performance on less frequent senses (LFS). \citet{mg-bert} jointly encode the context and multi-glosses of the target word to produce gloss embeddings that are almost orthogonal and emphasize the differences between glosses, thus achieving a better performance. 

Although there has been a lot of work on WSD recently which has improved performance and offered new insight on the problem, to the best of our knowledge there is no research trying to do data augmentation on context-gloss pairs and we think this is a notable research field to explore. 

Data augmentation is a important technique to increase model generality and prevent overfitting, so an increasing number of NLP data augmentation methods have been proposed in the recent years. One popular method is back translation \cite{back-translation} which leverages neural a translation model to generate paraphrases by translating from source to target language and back. Furthermore, \citet{EDA} showed that simply replacing words with synonyms is a straight-forward and useful data augmentation technique. Besides, \citet{CBERT} and \citet{kumar2020data} leverage pre-trained language model to augment sentences and consider contextual meaning at the same time. The above works suggest that textual data augmentation can play an important role in improving NLP task performance.

This paper brings the following contributions:
\begin{itemize}
    \item [1] Consider synset relationships (hypernym and hyponym) in WordNet to augment our training corpora.
    \item [2] Investigate the potential of different word substitution augmentation techniques applied to context-gloss pairs.
    \item [3] Explore the advantages in using back translation as an augmentation method for both gloss and context, showing that gloss back translation using German outperforms all other methods.
    \item [4] Analyze the insight gained by experimenting with several context-gloss augmentation techniques.
\end{itemize}

\section{Methods}

\begin{figure}[h]
    \centering
    \includegraphics[width=0.5\textwidth]{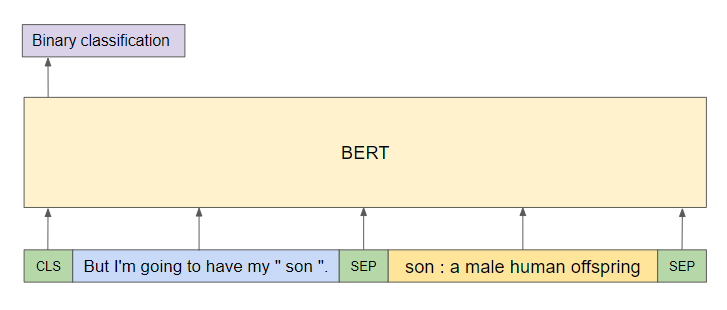}
    \caption{GlossBERT}
    \label{fig:my_label}
\end{figure}

\subsection{GlossBERT}
We use a pre-trained BERT model and fine-tune it using context-gloss pairs. 
As proposed by \citet{gloss-bert}, we use weak supervision on the context-gloss pairs. Context here refers to a sentence in the training corpora that contains the lemma we want to disambiguate.
The input to BERT is:
\begin{equation}\small[CLS]\;context\;[SEP]\;target\;:\;gloss\;[SEP]\end{equation}
As shown in Figure 1, we highlight the target word in the context by wrapping it with the special character \textbf{"}; we also prepend the target word itself to the gloss, separated by a \textbf{:} character.
For each lemma we want to disambiguate, we create \textit{n} records, one for each of the lemma's senses.
The label for each example is binary. If the gloss is the correct sense, then label it as 1; the mismatching senses are annotated as 0.
We then take the final hidden state of [CLS] as the representation of the whole sequence and add a binary classification layer. As a result, the model can learn to find the sense that is matched with the context among all each word's senses. GlossBERT is the baseline for our experiments, and all the following augmentation techniques aim to increase the number of context-gloss pairs used to train the model.

\subsection{Context-gloss augmentation}
We apply context-gloss augmentation because we believe that there are different ways to convey the same semantics and small variations in the sentence, such as paraphrasing and using synonyms, maintain the meaning of the sentence unchanged. We can use the increased number of records to prevent overfitting.

\subsubsection{Back translation}
Back translation augmentation is a well-known technique for generating paraphrases by translating sentences from a source language to a target language, and then translating back into the source language. Similarly to \citet{back-translation}, we increase the number of training examples by adding back translated context-gloss pairs. We use pre-trained models (see Appendix Table 4 for details) to perform back translation and experimented with German (De), Russian (Ru) and French (Fr) as target languages. 

\begin{table*}[t]
\centering
\begin{tabular}{  lrl }
\hline
\textbf{method} & \textbf{context} 
\\
\hline
- (baseline) & are you utilizing cafeteria space for company " meetings " or discussions ?  \\
MLM & are you utilizing cafeteria space \textcolor{red}{holding} company " meetings " or discussions? \\
synonym replacement & are you \textcolor{red}{use} cafeteria space for company " meetings " or discussions? \\
back translation & \textcolor{red}{do} you \textcolor{red}{use} the cafeteria for " meetings " or discussions \textcolor{red}{in companies}? \\
\hline
\textbf{method} & \textbf{gloss} \\
\hline
- (baseline) & a formally arranged gathering \\
back translation (De) &  a formally \textcolor{red}{established meeting} \\
hypernym & a group of persons together in one place \\
hypernym concatenation & a formally arranged gathering [SEP] a group of persons together in one place \\
\hline
\end{tabular}
\caption{
Different types of augmentation methods applied to an example context or gloss.
}
\end{table*}

\subsubsection{Hypernyms and hyponyms}
As suggested by \citet{glosses-wsd} we make use of the structured information contained in WordNet and retrieve the glosses for the senses' hypernyms and hyponyms - which is to say the concepts that represent respectively a generalisation or a specialisation of the given concept. We propose two approaches that make use of hypernyms' glosses. One is hypernym concatenation, which consists in appending a hypernym's gloss behind the original gloss separated by the [SEP] token. The second method consists in adding a new context-gloss pair for each original pair, with the new pair having the original gloss replaced by the hypernym's gloss.  

\subsubsection{Synonym replacement}
One popular NLP data augmentation technique is to replace the original word with a synonym. We leverage WordNet as the lexical knowledge base and substitute words with their synonyms. Each word has the probability \(p\) to be replaced, and we set \(p\) to 0.15. We only apply the synonym replacement method to context because on average the length of glosses is short, and substituting words in glosses can easily lead the glosses to significantly change meaning.

\begin{table*}[t]
\centering
\begin{tabular}{ lrl }
\hline
\textbf{Augmentation Method} & \textbf{Component} & \textbf{f1 score}
\\
\hline
- (baseline) & -  & 77.9\\
\hline
hypernyms & gloss & 79.3 \\
hypernyms concatenation & gloss & 79.6 \\
hypernyms and hyponyms &  gloss & 78.4\\
back translation (Ru) & gloss  & 78.8\\
back translation (Fr) & gloss  & 79.1 \\
back translation (De) & gloss  & \textbf{80.0}\\
back translation (Fr+De) & gloss  &  79.4\\
hypernyms concatenation + back translation (De) & gloss & 79.3 \\
hypernyms concatenation + back translation (Fr) & gloss & 79.1 \\
\hline
back translation (De) & context  & 78.9\\
synonym replacement &  context & 78.5\\
MLM & context  &  78.3\\
\hline
back translation (De) & gloss + context & 79.7\\
\hline
\end{tabular}
\caption{
The comparison of different context-gloss augmentation methods. We calculate testing f1 score based on ALL noun dataset.
}
\end{table*}

\subsubsection{Contextual word embeddings (MLM)}
 Language models like BERT have been pre-trained using a masked language model (MLM). We leverage this method and perform augmentation by masking random words in context sentence and substituting them with the model's prediction, taken from a list of \(k\) candidates. Here we set \(k\) to 3.

\section{Experiments}
\subsection{Training dataset and augmentation}
Following previous work \cite{glosses-wsd-2, glosses-wsd, gloss-bert} we choose SemCor3.0 as our basic training corpus, which is the largest corpus manually annotated with WordNet senses for WSD. We focus on the disambiguation of nouns only, exploring several data augmentation techniques on context-gloss pairs from WordNet.

\subsection{Evaluation datasets}
We evaluate our model on the benchmark datasets proposed by \citet{raganato2017word} which includes five standard English all-words fine-grained WSD datasets. They are Senseval-2 (\textbf{SE2}), Senseval-3 (\textbf{SE3}), SemEval-07 (\textbf{SE7}), SemEval-13 (\textbf{SE13}), and SemEval-15 (\textbf{SE15}). 
After filtering these datasets so that they contain noun only senses, we concatenate them to create a new dataset (\textbf{ALLn}) which we use to evaluate our models. 

\section{Results}
Table 2 shows the performance of different context-gloss augmentation methods on the ALLn evaluation datasets. 

All augmentation methods applied to gloss increase the performance.
Methods using hypernyms help performance a lot, as the f1 score increase from 0.779 to 0.793 and 0.796 depending on the method used. Whereas adding a hypernym's gloss as a new record might provide confounding signal that leads the model away from the original sense's meaning, the hypernym concatenation method achieves better performance since it provides extra information to the original sense. However, using both hypernyms and hyponyms can only improve f1 score to 0.784, demonstrating that hyponyms may be too generic and damage performance.

Remarkably, back translation offers an even more significant improvement. The f1 score increases to 0.788 when using Russian as a target language, to 0.791 when using French and to 0.800 when using German. 

As for context augmentation, back translation using German also takes the f1 score to 0.789, but the performance gain is not as significant when compared to gloss augmentation. As for the word-level augmentation methods, they both enhance the performance but not as much as back translation, since word substitution has a high probability of introducing grammatical errors. MLM performs worse than synonym replacement using WordNet, with an f1 score of 0.783 and 0.785 respectively. 

Even though doing back translation on gloss improves the performance considerably, the gain is reduced when back translation is used in combination with other methods, such as hypernym concatenation and back translation on context.  

\section{Discussion}
\subsection{Comparison among different translation languages}
We investigate German, French and Russian as target languages of back translation. German is one of the most closely related languages to English and like English belongs to the Germanic language family. Back translated sentences tend to have a word by word correspondence with the original, with most words being the same and some being replaced by a synonym. French has a considerable lexical overlap with English, but French belongs to the Romance language family and we can see there are some misleading statements in French's augmented sentences. Russian belongs to the Baltic-Slavic language family; the sentence patterns of English and Russian have greater dissimilarity and the augmentation noise is higher. We observe that the meaning of some of the augmented sentences changes a lot. A few examples of back translation by different languages are provided in the Appendix Table 3. 

Our results suggest that grammatical similarity has a greater impact than lexical similarity on the quality of translation for augmentation. Based on our experiment we think that WSD is sensitive to noise in semantic, so it's better to choose target languages that don't alter the sentence meaning too much.

\subsection{Augmentation on context versus gloss}
Table 2 shows that back translation (De) performs better when used on gloss compared to when it's used on context. There might be three reasons: firstly, glosses are generally shorter sentences, so the chance of a translation error distorting the meaning is lower.
Secondly, glosses have a more formal and standardised register when compared to context sentences, as the latter can originate from a varied array of sources (literary prose, newspaper article, etc.).
Lastly, context sentences often feature compound sentences and more complex syntactic relationship, whereas glosses usually contain straight-forward statements.

\section{Conclusion}
In this paper, we explore various types of data augmentation on context-gloss pairs data to improve WSD performance. This is the first research work that applies NLP data augmentation methods to WSD. We compare and analyze the results from different augmentation techniques. In our experiments, we find out that leveraging back translation on gloss only performs the best and improves the f1 score to 0.800. 
In the future we want to verify the generality of our findings by extending these methods to all-pos WSD and also to apply augmentation to different baseline models.

\bibliographystyle{acl_natbib}
\bibliography{anthology, acl2021}

\clearpage

\appendix

\begin{table*}
\centering
\begin{tabular}{ lrl }
\hline
\textbf{Language} & \textbf{Lemma} & \textbf{Back translated gloss}
\\
\hline
- (original) & face & the front of the human head from the forehead to the chin and ear to ear\\
De & face & the front of the human head from forehead to chin and from ear to ear \\
Fr & face & the forehead of the human head from the forehead to the chin and the ear to the ear \\
Ru & face & front of a person's head from head to chin and ear to ear \\
\hline
\hline
- (original) & day & time for Earth to make a complete rotation on its axis\\
De & day & time for the Earth to turn completely around its axis\\
Fr & day & it's time for Earth to make a complete rotation on its axis\\
Ru & day & the time when the Earth will complete a full rotation on its asteroid\\
\hline
\hline
- (original) & account & a short account of the news\\
De & account & a short report on the news\\
Fr & account & a brief account of the current situation\\
Ru & account & news at a glance\\
\hline
\end{tabular}
\caption{
Comparison of back translated glosses in German (De), French (Fr) and Russian (Ru).
}
\end{table*}

\begin{table*}
\centering
\begin{tabular}{  lrrr }
\hline
\textbf{target language} & \textbf{open-source package} & \textbf{target model} & \textbf{source model}
\\
\hline
german & nlpaug & transformer.wmt19.en-de & transformer.wmt19.de-en \\
russian & nlpaug & transformer.wmt19.en-ru & transformer.wmt19.ru-en \\
french & transformers  & Helsinki-NLP/opus-mt-en-fr & Helsinki-NLP/opus-mt-fr-en \\
\hline
\end{tabular}
\caption{
Open-source packages and models used for back translation
}
\end{table*}

\section{Experimental Setup}
We modify the maximum number of tokens input to BERT from 512 (GlossBERT usage) to 128 because we have checked that the more than 99 percent of context-gloss pairs have less than 128 tokens.
For the training hyper-parameters, we set the batch size to 96, learning rate to 2e-5 and the number of epochs to 4 for all the experiments.

\section{Error Analysis}
A deeper analysis of our results reveal some patterns worth reporting. We specifically look at our failed predictions and find there are some cases where the models fails to infer the correct sense and instead predicts a completely unrelated one.
However, in the remaining failed predictions, we notice that in a sizeable portion of instances the predicted sense is very similar to the true sense, often with the two senses sharing a very similar gloss. In some cases the similarity is so pronounced that the uniqueness or correctness of the true sense can be put into discussion. This suggests that WordNet might be too fine-grained in how it partitions lemmas into senses for any practical application of WSD.


\end{document}